\documentclass[conference]{IEEEtran}
\IEEEoverridecommandlockouts
\usepackage{cite}
\usepackage{amsmath,amssymb,amsfonts}
\usepackage{algorithmic}
\usepackage{graphicx}
\usepackage{textcomp}
\usepackage{epstopdf}
\usepackage[table, dvipsnames]{xcolor}
\begin{document}

\title{%Partial optimal matching and alignment between acoustic and linguistic knowledge transfer learning for ASR\\
	%Partial optimal matching between acoustic and symbolic tokens\\
	%Dual partial optimal matching \\
	%New insights on optimal alignment between acoustic and linguistic spaces in knowledge transfer learning for ASR\\
	New Insights into Optimal Alignment of Acoustic and Linguistic Representations for Knowledge Transfer in ASR \\
	%Revisiting Acoustic-Linguistic Alignment for Knowledge Transfer in Automatic Speech Recognition  
%\thanks{Acknowledgment: The work is partially supported by JSPS KAKENHI No. 24K15004.}
}

%Add authors here
\author{\IEEEauthorblockN{Xugang Lu, Peng Shen, Hisashi Kawai}\\
\IEEEauthorblockA{\textit{ National Institute of Information and Communications Technology, Japan}}\\
\\
%xugang.lu@nict.go.jp

}

\maketitle

\begin{abstract}

Aligning acoustic and linguistic representations is a central challenge to bridge the pre-trained models in knowledge transfer for automatic speech recognition (ASR). This alignment is inherently structured and asymmetric: while multiple consecutive acoustic frames typically correspond to a single linguistic token (many-to-one), certain acoustic transition regions may relate to multiple adjacent tokens (one-to-many). Moreover, acoustic sequences often include frames with no linguistic counterpart, such as background noise or silence may lead to imbalanced matching conditions. In this work, we take a new insight to regard alignment and matching as a detection problem, where the goal is to identify meaningful correspondences with high precision and recall ensuring full coverage of linguistic tokens while flexibly handling redundant or noisy acoustic frames in transferring linguistic knowledge for ASR. Based on this new insight, we propose an unbalanced optimal transport-based alignment model that explicitly handles distributional mismatch and structural asymmetries with soft and partial matching between acoustic and linguistic modalities. Our method ensures that every linguistic token is grounded in at least one acoustic observation, while allowing for flexible, probabilistic mappings from acoustic to linguistic units. We evaluate our proposed model with experiments on an CTC-based ASR system with a pre-trained language model for knowledge transfer. Experimental results demonstrate the effectiveness of our approach in flexibly controlling degree of matching and hence to improve ASR performance. \\

\end{abstract}


\section{Introduction}
Pre-trained language models (PLMs) have been leveraged to improve the performance of automatic speech recognition (ASR) with end-to-end (E2E) acoustic model frameworks \cite{Li2022,FNARBERT,NARBERT,KuboICASSP2022,Choi2022,Futami2022,Higuchi2023,CIFBERT1, wav2vecBERTSLT2022,CTCBERT1,CTCBERT2,DengICASSP2024,CIFBERT2}. Rather than stacking text encoders of PLMs on top of an acoustic encoder in ASR modeling \cite{NARBERT}, transferring linguistic knowledge encoded in the PLMs to acoustic encoding via cross-modal transfer learning or knowledge distillation (KD) has been studied \cite{KuboICASSP2022,FNARBERT,Futami2022,Choi2022,Higuchi2023,Cho2020,Cross2021}. With this cross-modal knowledge transfer learning, the rich context-dependent linguistic information from PLMs could be efficiently integrated in acoustic representation learning for better acoustic modeling, and the PLMs are not necessary to be involved in during inference stage which could enhance the parallel decoding for ASR \cite{CTCBERT1,CTCBERT2,CIFBERT1, CIFBERT2}.

Effective alignment and matching between acoustic and linguistic representations is a critical component of cross-modal knowledge transfer learning, largely due to the inherent modality gap between these two domains. Despite its significance, achieving robust alignment remains a challenging problem. The mapping between acoustic frames (or segments) and linguistic tokens is structurally asymmetric and non-uniform. Typically, multiple consecutive acoustic frames align with a single linguistic token (many-to-one), while in certain transitional regions, especially during rapid speech, an individual acoustic segment may correspond to multiple adjacent tokens (one-to-many). Additionally, acoustic sequences often contain redundant or non-informative frames, such as silence, background noise, or disfluent speech, which lack meaningful linguistic counterparts. These characteristics introduce substantial uncertainty and distributional imbalance, making standard alignment strategies often based on balanced, monotonic, or one-to-one assumptions insufficient. To address these challenges, we introduce a novel perspective that frames alignment and matching as a detection problem. Rather than enforcing rigid correspondences, the objective becomes identifying and matching informative acoustic frames to linguistic tokens while simultaneously rejecting irrelevant or noisy observations. This perspective naturally lends itself to the use of precision and recall metrics, as in classical detection tasks, aiming to maximize meaningful acoustic-linguistic alignments with high precision and recall while ensuring complete coverage of linguistic tokens and minimizing spurious associations.

Building on this new perspective, we propose a novel alignment framework grounded in unbalanced optimal transport (UOT) theory~\cite{VillanoBook, UOT}, which explicitly accounts for the distributional mismatch and structural asymmetry between acoustic and linguistic representations. In contrast to traditional alignment methods, our approach enables soft, partial, and inherently asymmetric matchings. It ensures comprehensive coverage of linguistic tokens while flexibly attenuating the influence of irrelevant or unmatched acoustic frames. Crucially, the framework guarantees that each linguistic unit is anchored to at least one meaningful acoustic observation, thereby supporting robust and semantically faithful alignment in cross-modal knowledge transfer.

To evaluate the proposed method, we integrate it into a CTC-based automatic speech recognition (ASR) system~\cite{CTCASR}, which is a popular E2E-ASR model framework~\cite{HierarchicalCTC, intermediateCTC}, incorporating a pre-trained language model to facilitate knowledge transfer. Experimental results show that our approach enables more effective alignment and leads to improved recognition performance, underscoring the advantage of detection-based alignment through UOT modeling in cross-modal transfer learning for ASR.

The remainder of this paper is structured as follows. Section \ref{sec:proposed} introduces the proposed approach, detailing the architecture of the cross-modal knowledge transfer model. We first present a novel perspective that frames alignment and matching as a detection problem, followed by the formulation of UOT, which serves as the core mechanism for flexible and robust cross-modal alignment. Section \ref{sec:exp} describes the experimental setup and provides a comprehensive evaluation of our method, including comparisons with baseline approaches. Finally, Section \ref{sec:conclusion} concludes the paper and outlines directions for future work.

\section{Proposed Method}
\label{sec:proposed}
\begin{figure}[ht]
	\centering	
	\includegraphics[width=8cm, height=6cm]{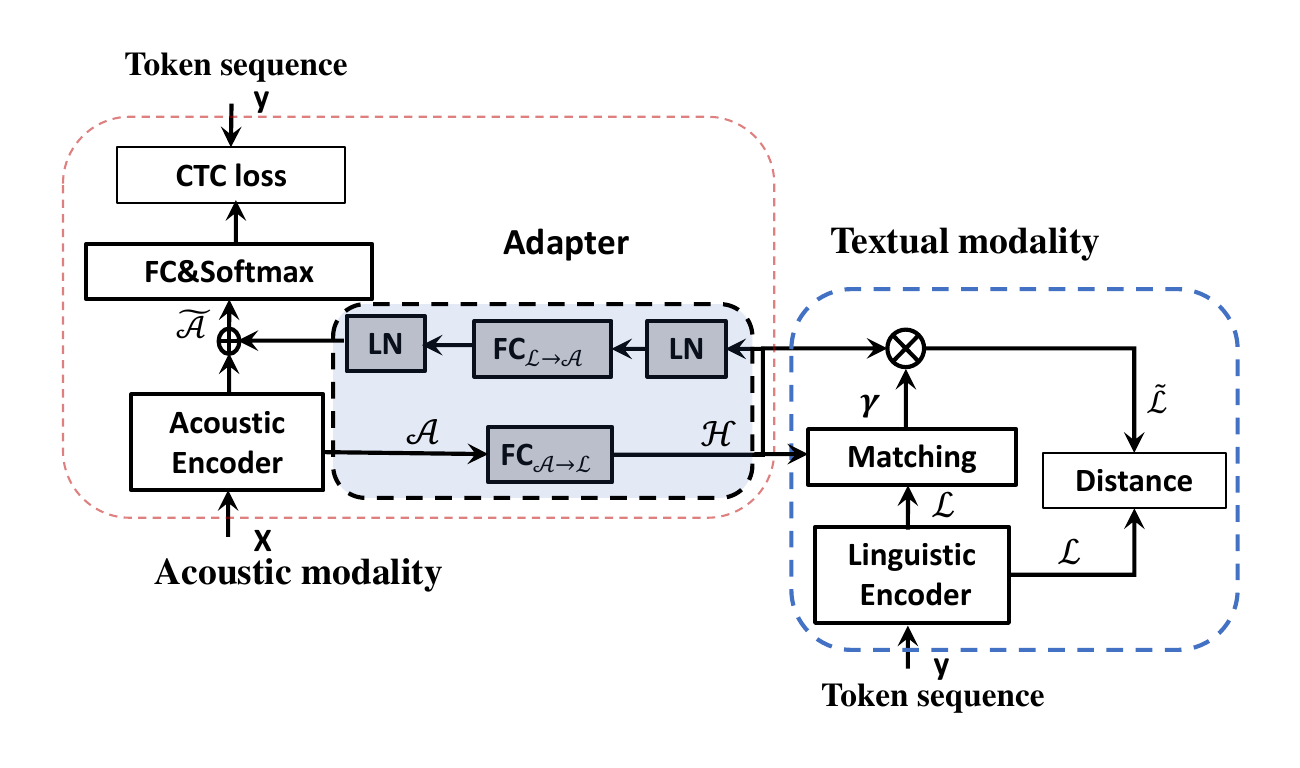}
	\caption{The proposed cross-modal knowledge transfer learning framework for ASR.}
	%\vspace{-2mm}		
	\label{fig:fig1}
\end{figure}

The proposed model framework is showed in Fig. \ref{fig:fig1} which is inspired by the study in \cite{ASRU2023Lu}. In this framework, two encoders, i.e., acoustic and linguistic encoders are used for exploring acoustic and linguistic feature representations $\mathcal{A}$ and $\mathcal{L}$ as: 
\begin{equation}
	\begin{array}{l}
		\begin{aligned}
	\mathcal{A} &= {\rm Encoder}_{\mathcal{A}} ({\bf X}) \\	
	\mathcal{L} &= {\rm Encoder}_{\mathcal{L}} ({\bf y}),	
\end{aligned}
\end{array}
\end{equation}
where ${\bf X}$ and ${\bf y}$ are acoustic and linguistic inputs, ${\rm Encoder}_{\mathcal{A}}$ and ${\rm Encoder}_{\mathcal{L}}$ denote the transforms of the two encoders, respectively. Between the two encoders, there is an `Adapter' module for feature transforms during knowledge transfer learning. In right branch of Fig. \ref{fig:fig1}, a matching module is designed which serves as the key in cross-modal knowledge transfer learning. The function of this matching module is to align and match acoustic and linguistic representations for efficient transfer learning. 
Before giving details of the model, we first explain our new insights on the alignment and matching problem.  
\subsection{Cross-modal matching: A Detection point of view}
\begin{figure}[tb]
	\centering	
	\includegraphics[width=8cm, height=8cm]{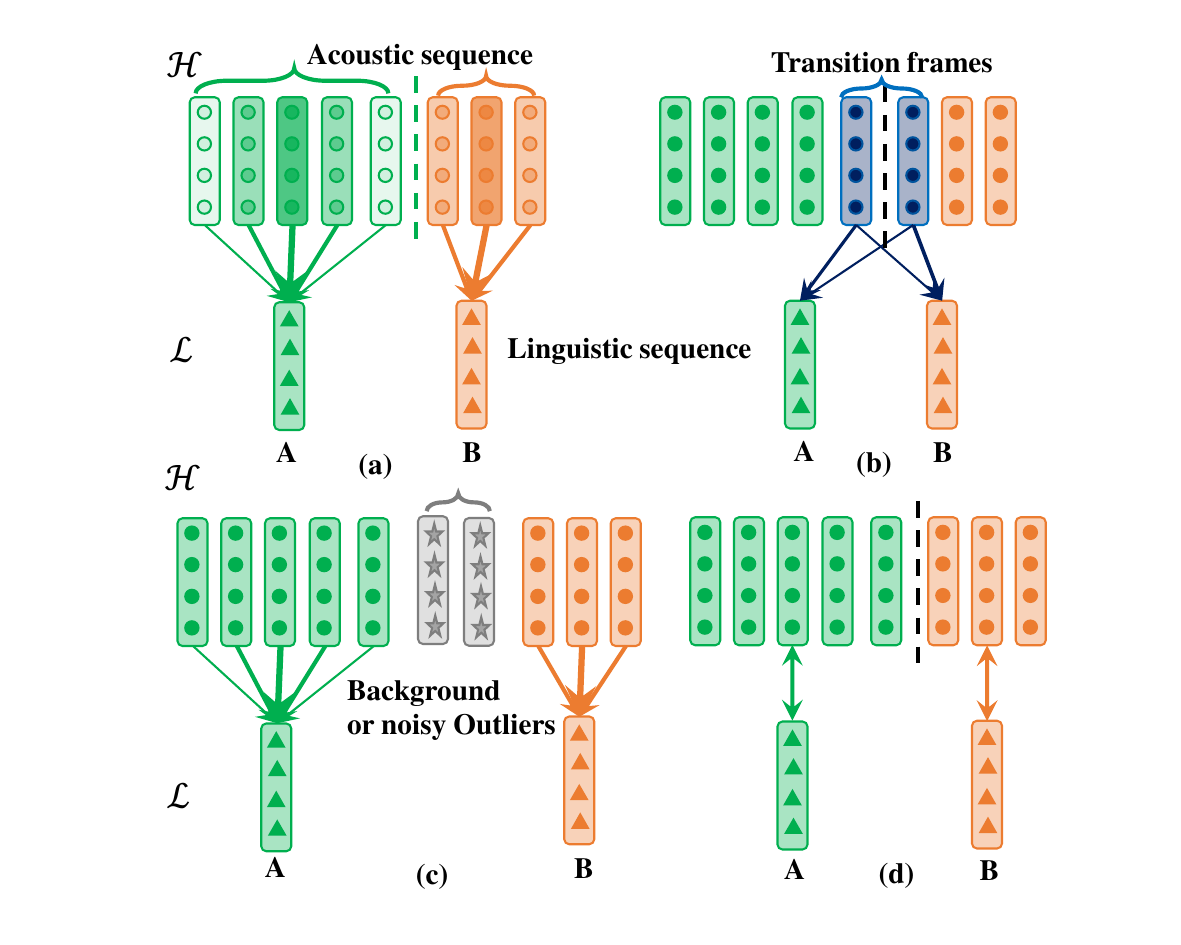}
	\caption{Alignment and matching between acoustic and linguistic representations (thickness of arrow lines represent degree of cross-modal associations): (a) several consecutive acoustic frames are matched to one token (many to one matching); (b) acoustic transition frames are associated to two different tokens (one to many matching); (c) acoustic background or outliers has no corresponding linguistic tokens (NULL matching); (d) each linguistic token should have at least one best match in acoustic space.}
	%\vspace{-2mm}		
	\label{fig:fig2}
\end{figure}

In the cross-modalities, for a paired speech-text sequences $\mathcal{A}$ and $\mathcal{L}$, suppose the acoustic sequence is represented as:
\[
\mathcal{A} = \{ \mathbf{a}_1, \dots, \mathbf{a}_m \}, 
\]
where $\mathbf{a}_i \in \mathbb{R}^{d_a},$ is acoustic representation vector with dimension $d_a$ extracted from acoustic encoder, and the linguistic sequence as:
\[
\mathcal{L} = \{ \mathbf{l}_1, \dots, \mathbf{l}_n \}, 
\]
where $ \mathbf{l}_j \in \mathbb{R}^{d_l}$ is token representation vector with dimension $d_l$ explored from linguistic encoder. Since feature dimension $d_a$ of acoustic representation is different from that of linguistic representation $d_l$, in the first step of sequential matching, a dimension matching transform is applied on acoustic representation as: 
\begin{equation}
	{\bf h}_i  = {\rm FC}_{\mathcal{A} \to \mathcal{L}} {\rm (}{\bf a}_i ),
\end{equation}
where $\mathbf{h}_i \in \mathbb{R}^{d_l}$, hence dimension of acoustic representation $\mathcal{H} = \{ \mathbf{h}_1, \dots, \mathbf{h}_m \}$ and linguistic representation $\mathcal{L}$ is matched. Moreover, in matching between acoustic and linguistic sequences, usually $m>n$, i.e., the number of acoustic frames is much larger than the number of linguistic tokens. Correspondingly, in alignment and matching, several situations should be fulfilled, for example, multiple consecutive acoustic frames could correspond to a single linguistic token (many-to-one), while occasionally, certain acoustic transitions may align with adjacent tokens (one-to-many), moreover, redundant or irrelevant acoustic features which do not correspond to any linguistic unit should be discarded. This necessitates a soft, selective, and asymmetric alignment model. We adopt a new perspective on this problem by framing feature alignment and matching as a detection task. From this viewpoint, alignment aims to identify meaningful and reliable correspondences between the acoustic and linguistic feature representations with high precision and recall in order to effectively matching relevant acoustic frames to linguistic tokens while rejecting spurious or noisy associations and ensuring complete coverage of all linguistic tokens. A more detailed illustration of this alignment and matching process is provided in Fig. \ref{fig:fig2}.

In Fig. \ref{fig:fig2}, acoustic sequence vectors $\mathcal{H}$ are represented with circles, and linguistic sequence vectors $\mathcal{L}$ are denoted with triangles (only two token representations as `A' and `B' are showed). When matching between acoustic representations and linguistic tokens, as illustrated in Fig. \ref{fig:fig2}-a, in most cases, it is a many to one matching problem, i.e., several consecutive frames are aligned and matched to one token with different degree of associations indicated as different thicknesses of arrows. In some occasions, it is difficult to definitely assign acoustic frames in transition segments to one token, i.e., one acoustic frame may be associated to two neighboring tokens with different uncertainty which is a one to many problem as denoted in Fig. \ref{fig:fig2}-b. Moreover, in linguistic representations, redundant acoustic frames or outliers may have no corresponding transcriptions as shown in Fig. \ref{fig:fig2}-c, in this scenario, NULL matching or alignment occurs. Furthermore, in associating linguistic tokens to acoustic correspondences, we need to guarantee that each token should be aligned to at least one best acoustic frame in matching as showed in Fig. \ref{fig:fig2}-d. Based on this analysis, in alignment and matching between acoustic and linguistic representations, we need to keep high matching precision and low missing matches which corresponding to low false positives and high recall in signal detection task. This requirement fits well to one mathematical theory-UOT for cross-modal transform associations, i.e., controlling the marginal distributions cross modalities during optimal transport.  

\subsection{Unbalanced Optimal Transport Formulation (UOT)}
To model flexible and noisy matching in cross-modal transfer learning, we adopt the UOT framework. This framework allows for partial mass transport between two distributions and smooths the transport plan to avoid overconfident alignments when taking entropy regularization of transport plan into consideration \cite{UOT}. Suppose acoustic and linguistic representations are sampled from two discrete probability distributions or measures $\mathcal{H} \sim \mu$ and $\mathcal{L} \sim \nu$ with:
\begin{equation}
	{\mu} = \sum_{i=1}^{m} w_i \delta_{\mathbf{h}_i}, \quad {\nu} = \sum_{j=1}^{n} v_j \delta_{\mathbf{l}_j},
\end{equation}
%\begin{align*}
%	{\mu} &= \sum_{i=1}^{m} w_i \delta_{\mathbf{h}_i} \\
%	{\nu} &= \sum_{j=1}^{n} v_j \delta_{\mathbf{l}_j},
%\end{align*}
where \( \delta_{\mathbf{x}} \) denotes the Dirac delta at \( \mathbf{x} \), \( w_i \) and \( v_j \) are non-negative weights for acoustic and linguistic samples as in two marginal probability distribution vectors as:
\begin{equation}
\begin{array}{l}
	\begin{aligned}
\mathbf{w} &= (w_1 ,w_2 ,...,w_m )^\top \in \mathbb{R}^{m} \\ 
\mathbf{v} &= (v_1 ,v_2 ,...,v_n )^\top \in \mathbb{R}^{n}, \\
    \end{aligned}
\end{array}
\end{equation}

Let \( \mathbf{C} \in \mathbb{R}^{m \times n} \) be the cost matrix with entries \( C_{ij} = c(\mathbf{h}_i, \mathbf{l}_j) \) (distance function of two compared vectors as $c(.)$), and \( \gamma \in \mathbb{R}_{+}^{m \times n} \) denotes a transport plan, where \( \gamma_{ij} \) is the mass moved from \( \mathbf{h}_i \) to \( \mathbf{l}_j \). The entropy-regularized UOT problem is formulated as:
\begin{equation}
	%\begin{array}{l}
	%	\begin{aligned}
	L_{\rm UOT} \mathop  = \limits^\Delta\min_{\gamma \in \mathbb{R}_{+}^{m \times n}} 
	\sum_{i,j}^{m,n}  \gamma_{ij} C_{ij}
	+ L(\mathbf{w}, \mathbf{v})
	- \varepsilon \cdot H(\gamma),
%\end{aligned}
%\end{array}
	\label{eq:uot-entropy}
\end{equation}
with $L(\mathbf{w}, \mathbf{v})$, a penalty function defined as:
\begin{equation}
	L(\mathbf{w}, \mathbf{v}) \mathop  = \limits^\Delta \lambda_1 \, \mathcal{D}( \gamma \mathbf{1}_n \,\|\, \mathbf{w})
	+ \lambda_2 \, \mathcal{D}( \gamma^\top \mathbf{1}_m \,\|\, \mathbf{v}),
	\label{eq:penalty}
\end{equation}
where
\begin{itemize}
	\item \( \mathbf{1}_m \in \mathbb{R}^{m} \) and \( \mathbf{1}_n \in \mathbb{R}^{n} \) are vectors of ones;
	\item \( \gamma \mathbf{1}_n \in \mathbb{R}^{m} \) and \( \gamma^\top \mathbf{1}_m \in \mathbb{R}^{n} \) are the row and column marginals of \( \gamma \);
	\item \( \mathcal{D}(\cdot \| \cdot) \) is a Kullback--Leibler divergence based distance function;
	\item \( \lambda_1, \lambda_2 \geq 0 \) control the penalty on deviation from the two original marginals;
	\item \( H(\gamma) = - \sum_{i,j} \gamma_{ij}  \log \gamma_{ij} \) is the entropy of the transport plan;
	\item \( \varepsilon > 0 \) is the entropy regularization coefficient.
\end{itemize}

The entropy term encourages smooth, diffused alignments rather than hard assignments. This is beneficial in the context of noisy or ambiguous acoustic frames and facilitates fast approximation using iterative Sinkhorn-type algorithms with a solution as \cite{Cuturi2013}:
\begin{equation}
\gamma^* = \operatorname{diag}(\boldsymbol\alpha) K \operatorname{diag}(\boldsymbol\beta),
\end{equation}
where \( \boldsymbol\alpha \in \mathbb{R}^m \) and \( \boldsymbol\beta \in \mathbb{R}^n \) are the scaling vectors, and $K$ is the Gibbs kernel matrix with elements defined as:
\begin{equation}	
	K_{ij} = \exp\left( - \frac{C_{ij}}{\varepsilon} \right)	
\end{equation}
The iteration process to update the scaling factors for solving UOT is defined as:
\begin{equation}
	\begin{array}{l}
		\begin{aligned}
%\begin{align*}
	\boldsymbol\alpha^{(t+1)} &= \left( \frac{\mathbf{w}}{K \boldsymbol\beta^{(t)}} \right)^{\frac{\lambda_1}{\lambda_1 + \varepsilon}} \\
	\boldsymbol\beta^{(t+1)} &= \left( \frac{\mathbf{v}}{K^\top \boldsymbol\alpha^{(t+1)}} \right)^{\frac{\lambda_2}{\lambda_2 + \varepsilon}}.
%\end{align*}
\end{aligned}
\end{array}
\label{eq:scaling}
\end{equation}
These updates are repeated until convergence or stopped when reaching a given fixed threshold. 
\subsection{Alignment and matching via marginal control}
From Eqs. (\ref{eq:uot-entropy}), (\ref{eq:penalty}) and (\ref{eq:scaling}), we can see that when \( \lambda_1, \lambda_2 \gg \varepsilon \), the original marginals are enforced strictly which is a common case of the standard (balanced) entropy-regularized optimal transport \cite{Cuturi2013}. When \( \lambda_1 = \lambda_2 = 0 \), the problem reduces to a free matching under entropy-regularized cost. Also from Eq. (\ref{eq:scaling}), we can see that the scaling $\boldsymbol\alpha^{(t+1)}$ and $\boldsymbol\beta^{(t+1)}$ are row and column normalization \cite{Cuturi2013,ICASSP2024Lu}, i.e., dual normalization for matching on both sides of cross modalities. With setting of different $\lambda_1$ and $\lambda_2$, we can control and enable efficient computation of soft and partial alignments to fulfil different degrees of precision and recall as a detection task, i.e., the flexibility of UOT allows directional control over the alignment behavior through appropriate settings of \( \lambda_1 \) and \( \lambda_2 \):
\begin{itemize}
	\item Acoustic-to-Linguistic (A2L) alignment: To ensure that every linguistic token is aligned (high recall), we set \( \lambda_2 > \lambda_1 \). This forces mass coverage over linguistic units while permitting selective skipping or discarding of noisy and outlier acoustic frames including NULL acoustic matching.
	
	\item Linguistic-to-Acoustic (L2A) alignment: To account for as much of the acoustic input as possible (high precision), we set \( \lambda_1 > \lambda_2 \), ensuring acoustic frames are matched even if some linguistic tokens are less activated.
\end{itemize}
This matching strategy in model training tries to encourage bidirectional consistency, and the resulting optimal transport plan \( \gamma^\ast \) represents a soft alignment matrix that assigns probabilistic mass between acoustic and linguistic elements, effectively grounding linguistic tokens in observed speech while avoiding overfitting to background noisy frames in transfer learning.

\subsection{Objective functions in knowledge transfer learning}
In cross-modal alignment, the acoustic feature can be projected onto the linguistic space based on the estimated transport coupling as:
\begin{equation}
	{\widetilde {\mathcal{L}}}_{\mathcal{L} \leftarrow  \mathcal{H}} = {\gamma ^*}^\top  \times \mathcal{H}  \in \mathbb{R}^{n  \times d_l },  
\end{equation}
with element $\widetilde {\mathcal{L}}_{\mathcal{L} \leftarrow  \mathcal{H}} = \{ {\widetilde {\mathbf{l}}_1}, \dots, {\widetilde{\mathbf{l}}_n} \}$, where $\gamma ^*$ is the optimal transport coupling matrix based on Eqs. (\ref{eq:uot-entropy}), (\ref{eq:penalty}) and (\ref{eq:scaling}). Subsequently, the alignment loss is defined as:
\vspace{-2mm}
\begin{equation}
	L_{{\rm align}}  \mathop  = \limits^\Delta \sum\limits_{j = 1}^{n} {1 - \cos \left( {{\bf \tilde l}_j ,{\bf l}_j } \right)} 
	\label{eq:Align}
\end{equation}
For efficient linguistic knowledge transfer to acoustic encoding, the following transform is designed as indicated in Fig. \ref{fig:fig1}:
\begin{equation}			
	\widetilde {\mathcal{A}}  = \mathcal{A}  + {\rm LN(FC}_{\mathcal{L} \to \mathcal{A}} {\rm (LN(}{\mathcal{H}} {\rm )))} \in \mathbb{R}^{m  \times d_a }, 	
	\label{eq:adapter}
\end{equation}
Based on this new representation $\widetilde {\mathcal{A}}$, the probability prediction for ASR is formulated as ${\bf \tilde P} = {\rm Softmax}( {{\rm FC}(\widetilde {\mathcal{A}})} )$,
%\begin{equation}
%{\bf \tilde P} = {\rm Softmax}\left( {{\rm FC1}\left({\bf H}^{\rm AL} \right)} \right),
%\label{eq:softmaxadd}
%\end{equation}
where `FC' is a full-connected linear transform with output size the same as that of vocabulary tokens. Finally, the total loss in model training is defined as:
%\vspace{-2mm}
\begin{equation}
	L\mathop  = \limits^\Delta  \eta L_{{\rm CTC}} ({\bf \tilde P},{\bf y} ) + (1 - \eta ){(L_{{\rm align}}  + L_{{\rm UOT}} )},   
	\label{eq:totalloss} 	
\end{equation}
where $L_{{\rm CTC}} ({\bf \tilde P},{\bf y} )$ is CTC loss, ${L_{{\rm align}}}$ and ${L_{{\rm UOT}} }$ are cross-modality alignment loss and UOT loss, respectively. After the model is trained, only the left branch of Fig. \ref{fig:fig1} is retained for ASR inference.  

\section{Experiments and Results}
\label{sec:exp}
ASR experiments were conducted on an open-source Mandarin speech corpus AISHELL-1 \cite{AISHELL1} to evaluate the proposed algorithm. The data corpus comprises three datasets: a training set with 340 speakers (150 hours), a development (or validation) set with 40 speakers (10 hours), and a test set with 20 speakers (5 hours). Data augmentation as used in \cite{AISHELL1} was applied. Given the tonal nature of the Mandarin language in the ASR task, in addition to using 80-dimensional log Mel-filter bank features, three extra acoustic features related to fundamental frequency, i.e.,  F0, delta F0 and delta delta F0, were utilized as raw input features. These features were extracted with a 25ms window size and a 10ms shift. 
\subsection{Model configurations}
In Fig. \ref{fig:fig1}, acoustic and linguistic encoders can be any type of model with neural network-based architectures. For easy control of our experiments, a conformer-based acoustic encoder and transformer-based linguistic encoder are adopted the same as used in \cite{ASRU2023Lu}. The acoustic encoder consists of CNN-based subsampling blocks (two CNN blocks with 256 channels, kernel size 3, stride 2, and ReLU activation function in each), and $16$ layers of conformer blocks \cite{conformer2020} with each having a kernel size of 15, attention dimension $d_a=256$, $4$ attention heads, and a 2048-dimensional feed-forward layer. Before cross-modal knowledge transfer learning, the acoustic encoder is pre-trained with CTC loss for ASR task. The pretrained LM `bert-base-chinese' from huggingface is used as the linguistic encoder during knowledge transfer learning \cite{Huggingface}, where $12$ transformer encoders of BERT \cite{BERT} are applied with the token size of 21128, and the dimension of linguistic feature representation is $d_l=768$. Correspondingly, in the adapter module, ${\rm FC}_{\mathcal{A} \to \mathcal{L}}$ and ${\rm FC}_{\mathcal{L} \to \mathcal{A}}$ are full connected linear transforms with $256*768$ and $768*256$ transform matrices, respectively. 
\subsection{Setting of hyper-parameters}
The proposed learning algorithm involves several hyperparameters, each of which can influence the final ASR performance in the cross-modal transfer learning settings. We will specify their values when presenting the experimental results. In our experiments, the loss balancing coefficient $\eta$ in Eq.~(\ref{eq:totalloss}) is fixed at 0.3. For optimization, we use the Adam optimizer \cite{Adam} with an initial learning rate of 0.002, following a learning rate schedule that includes 20,000 warm-up steps. The model with cross-modal transfer is trained for 130 epochs, and the final model used for evaluation is obtained by averaging the checkpoints from the last 10 epochs.
\subsection{Feature alignment and matching based on UOT}
As shown in Eqs. (\ref{eq:uot-entropy}) and (\ref{eq:penalty}), UOT relaxes the marginal constraints using divergence penalties, allowing for mass mismatch during transport. This property enables UOT to flexibly model the many-to-one and one-to-many relationships commonly found in acoustic-to-linguistic correspondences. Such flexibility allows the alignment process to focus on semantically meaningful associations while ignoring spurious or redundant frames, resulting in more robust and faithful feature matching.

\begin{figure}[tb]
	\centering	
	\includegraphics[width=8cm, height=7cm]{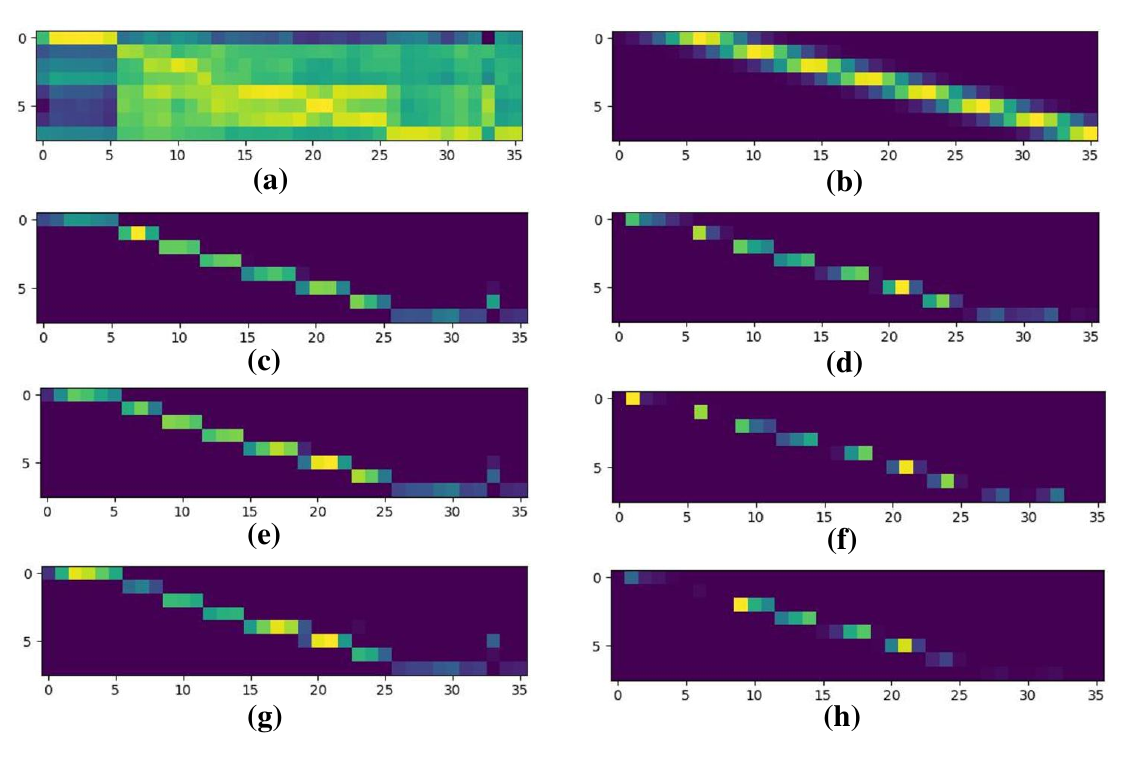}
	\caption{Optimal transport coupling with different weights in controlling of the marginal distributions in alignment and matching between acoustic and linguistic representations: (a) cosine similarity matrix between acoustic and linguistic representations; (b) uniform alignment and matching with Gaussian-shaped temporal coherence (acoustic sequence is uniformly segmented and matched to the underlying tokens); (c) $\lambda_1=10.0$, $\lambda_2=10.0$; (d) $\lambda_1=0.1$, $\lambda_2=1.0$; (e) $\lambda_1=1.0$, $\lambda_2=1.0$; (f) $\lambda_1=0.01$, $\lambda_2=1.0$; (g) $\lambda_1=1.0$, $\lambda_2=0.01$; (h) $\lambda_1=0.05$, $\lambda_2=0.05$.}
	%\vspace{-2mm}		
	\label{fig:fig3}
\end{figure}

Fig. \ref{fig:fig3} illustrates the effect of varying the marginal penalty parameters $\lambda_1$ and $\lambda_2$ on alignment behavior in cross-modal transfer learning. Fig. \ref{fig:fig3}-(a) shows the cosine similarity matrix between normalized acoustic and linguistic representations (pre-trained models). Fig. \ref{fig:fig3}-(b) displays a uniform segmentation of the acoustic sequence with Gaussian-shaped alignment to linguistic tokens. However, since acoustic durations vary non-uniformly across tokens, realistic alignments are typically sparse and uneven. Figs. \ref{fig:fig3}-(c), (d), (e), (f), (g), (h) show the transport coupling representing the alignment and matching between the two modalities with different settings of marginal controlling parameters $\lambda_1$ and $\lambda_2$. In these figures, horizontal axis is acoustic frame index, and vertical axis denotes linguistic token index, and the entropy regularization parameter in Eq. (\ref{eq:uot-entropy}) is fixed with $\epsilon=0.05$. When both $\lambda_1$ and $\lambda_2$ are large, e.g., in Fig. \ref{fig:fig3}-(c), most acoustic frames and linguistic tokens are retained in the alignment, though with varying coupling intensities. Conversely, when both values are small, e.g., in Fig. \ref{fig:fig3}-(h), the alignment becomes more selective, discarding many acoustic frames and even some linguistic tokens. Figs. \ref{fig:fig3}-(d), (e), (f), (g) with different values and combinations demonstrate how adjusting $\lambda_1$ and $\lambda_2$ controls the sparsity of the alignment: smaller values lead to stricter filtering, while larger values preserve broader correspondences. These results align with the interpretation of alignment as a detection problem, where precision and recall are key performance metrics. Overall, the UOT framework enables flexible control over cross-modal feature alignment, allowing the model to emphasize essential knowledge transfer while down-weighting redundant or noisy acoustic frames.      

\subsection{Results}
\label{sec:results}
Based on the above analysis, ASR experiments are carried out with different model parameter settings. In our experiments, only the acoustic modeling branch with the adapter blocks were kept so that to maintain the decoding speed similar to that of the CTC-based decoding (only the CTC greedy search-based decoding strategy was used in this paper), the results are showed in table \ref{tab:tab1}. The performance of several baseline systems were also given in table \ref{tab:tab1} for comparison. In table \ref{tab:tab1}, baseline models Conformer+CTC \cite{wenet2.0}, Conformer+CTC/AED \cite{Kim2017, Hori2017,Watanabe2017}, and linguistic knowledge transfer learning-based model NAR-BERT-ASR \cite{NARBERT} (stacking BERT model on acoustic encoder) are our own implementations which results were much better than those of the original results \cite{Kim2017, Hori2017,Watanabe2017,NARBERT}. In all experiments of table \ref{tab:tab1}, the weighting coefficient in Eq. (\ref{eq:totalloss}) is fixed with $\eta=0.3$. The entropy regularization controls the smoothness and sparseness of the coupling matrix and hence the knowledge transfer, we empirically examined the effect on ASR performance with different parameters, when $\epsilon$ is too small (in our case less than $0.02$), we could not obtain stable solutions. And when it is too large, the performance starts to decrease due to the large diffused matching. Moreover, from Eq. (\ref{eq:scaling}), we can see that the UOT is jointly controlled by combinations of $\epsilon$, $\lambda_1$ and $\lambda_2$. For easy control, in our experiments, $\epsilon=0.05$ is also fixed. With setting different marginal control parameters, the results are showed as `UOT-BERT-CTC' rows in table \ref{tab:tab1}. From this table, we can observe that in all these parameter settings, our proposed UOT-based transfer learning outperforms other compared systems. Moreover, when the weighting marginal parameters are large enough, e.g., $\lambda_1=10.0$, $\lambda_2=10.0$, theoretically, the alignment and matching is approximating to the study in \cite{ASRU2023Lu}, and the performance is almost comparable which confirms our proposal.   
\begin{table}[tb]
	\centering
	\caption{ASR performance on the AISHELL-1 corpus, CER (\%).}
	\vspace{-2mm}
	\begin{tabular}{|c||c||c|}
		\hline
		Methods &dev set &test set\\
		\hline		
		\hline		
		\rowcolor{lightgray}
		Conformer+CTC (Baseline)  &5.16 &5.76 \\		
		\hline	
		\rowcolor{lightgray}
		Conformer+CTC/AED (\cite{Watanabe2017,wenet2.0})  &4.31 &4.82 \\						
		\hline
		\rowcolor{lightgray}
		NAR-BERT-ASR (\cite{NARBERT}) &4.18 &4.68 \\
		\hline
		\rowcolor{lightgray}
		OT-BERT-CTC (\cite{ASRU2023Lu}) &3.81 &4.19 \\
		\hline
		\hline
		%\rowcolor{lightgray}		
		UOT-BERT-CTC ($\lambda_1=10.0$, $\lambda_2=10.0$)  &3.82 &4.21 \\				
		\hline
		%\rowcolor{lightgray}
		UOT-BERT-CTC ($\lambda_1=1.0$, $\lambda_2=1.0$) &3.70 &4.13 \\
		\hline
		%\rowcolor{lightgray}
		UOT-BERT-CTC ($\lambda_1=0.5$, $\lambda_2=1.0$) &\textbf{3.64} &\textbf{4.06} \\
		\hline
		%\rowcolor{lightgray}
		UOT-BERT-CTC ($\lambda_1=1.0$, $\lambda_2=0.5$) &3.81 &4.13 \\
		\hline						
	\end{tabular}
	%\vspace{-2mm}
	\label{tab:tab1}
\end{table}

\subsection{Discussion}
\begin{figure}[ht]
	\centering
	\includegraphics[width=8cm, height=3.5cm]{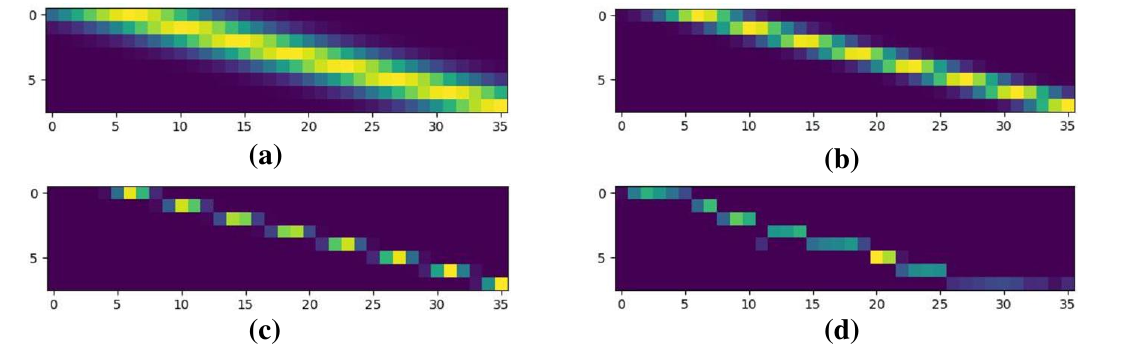}
	\caption{Gaussian-shaped uniform alignment vs adaptive alignment based on UOT: uniform alignment with Gaussian smooth window size of $10$ (a), window size of $5$ (b), window size of $2$ (c), and UOT with marginal control $\lambda_1=0.5$, $\lambda_2=0.5$ (d).}
	%\vspace{-2mm}		
	\label{fig:fig4}
\end{figure}

We further conducted experiments under various unbalanced conditions by adjusting the marginal control parameters in the UOT framework. The results are summarized in Table~\ref{tab:tab2}. For intuitive understanding, illustrative examples of alignment and matching under different configurations are shown in Fig.~\ref{fig:fig4}, where uniform alignment is compared with UOT-based alignment. From these results, we observe that while uniform alignment can transfer linguistic knowledge and improve ASR performance compared to a baseline CTC system, they often mix correct and incorrect matches, limiting overall alignment precision. This remains true whether all features are retained or a fixed proportion is discarded during alignment. In contrast, the proposed UOT-based approach provides adaptive control over the alignment and matching process, leading to more reliable performance gains. The results also highlight the importance of selecting appropriate penalty weights $\lambda_1$ and $\lambda_2$ to achieve optimal ASR results. While smaller acoustic marginal penalties help suppress the influence of redundant or noisy frames, the inherent disparity that acoustic sequences typically have more frames than corresponding linguistic tokens requires careful balancing of marginal penalties within the UOT framework to ensure effective and meaningful cross-modal alignment.
  
\begin{table}[tb]
	\centering
	\caption{Effect of marginal control in UOT on the ASR performance, CER (\%).}
	\begin{tabular}{|c||c||c|}
		\hline
		Methods &dev set &test set\\
		\hline
		\hline		
		\rowcolor{lightgray}
		Uniform alignment (window size = 10)  &4.31 &4.89 \\
		\hline
		\rowcolor{lightgray}
		Uniform alignment (window size = 5)  &4.33 &4.72 \\
		\hline
		\rowcolor{lightgray}
		Uniform alignment (window size = 2) &4.38 &4.81 \\
		\hline
		\hline
		%\rowcolor{lightgray}
		UOT-BERT-CTC ($\lambda_1=0.5$, $\lambda_2=0.5$) &3.73 &4.13 \\
		\hline
		%\rowcolor{lightgray}
		UOT-BERT-CTC ($\lambda_1=0.5$, $\lambda_2=0.1$) &3.88 &4.24 \\
		\hline	
		%\rowcolor{lightgray}	
		UOT-BERT-CTC ($\lambda_1=0.1$, $\lambda_2=0.5$)  &3.99 &4.44 \\		
		\hline
		%\rowcolor{lightgray}
		UOT-BERT-CTC ($\lambda_1=0.1$, $\lambda_2=0.1$) &4.00 &4.48 \\
		\hline
		%\rowcolor{lightgray}
		UOT-BERT-CTC ($\lambda_1=0.05$, $\lambda_2=0.05$) &4.08 &4.52 \\
		\hline
		%\hline
	\end{tabular}
	%\vspace{-3mm}
	\label{tab:tab2}
\end{table}

\section{Conclusion and future work}
\label{sec:conclusion}
This paper presents a novel perspective on the alignment between acoustic and linguistic representations in knowledge transfer for ASR. By recasting alignment as a detection problem, we highlight the importance of achieving high precision and recall in establishing meaningful correspondences, particularly in the presence of structural asymmetry and distributional mismatch across modalities. To address these challenges, we propose an UOT framework tailored for cross-modal knowledge transfer, which enables soft, partial, and probabilistic alignments that reflect the inherent structure of acoustic-to-linguistic mappings. By regulating the marginal constraints via divergence penalties, the UOT model flexibly matches partial distributions, ensuring that each linguistic token is grounded in at least one acoustic observation, while tolerating redundant or noisy acoustic frames without enforcing spurious matches. Experiments on a CTC-based ASR system with a pre-trained language model demonstrate that our method effectively controls alignment quality, leading to improved recognition performance in transfer learning settings. These results suggest that detection-based alignment grounded in unbalanced transport theory offers a principled and adaptable approach for bridging acoustic and linguistic modalities in ASR.

In this paper, model parameters were set empirically in our experiments. As shown in Eqs.~(\ref{eq:uot-entropy}), (\ref{eq:penalty}), and (\ref{eq:scaling}), these parameters influence the robustness and stability of the UOT formulation and should ideally be tuned jointly. In future work, beyond evaluating our proposed framework on additional tasks involving cross-modal alignment and matching, we also plan to investigate adaptive regularization strategies to further improve alignment robustness and generalization.

%\vspace{12pt}
%\color{red}
%IEEE conference templates contain guidance text for composing and formatting conference papers. Please ensure that all template text is removed from your conference paper prior to submission to %the conference. Failure to remove the template text from your paper may result in your paper not being published.

\end{document}